%% file: main.tex
\documentclass[10pt,twocolumn,letterpaper]{article}

\usepackage{wacv}      


\definecolor{wacvblue}{rgb}{0.21,0.49,0.74}
\usepackage[pagebackref,breaklinks,colorlinks,allcolors=wacvblue]{hyperref}

\usepackage{graphicx}
\usepackage{amsmath}
\usepackage{amssymb}
\usepackage{booktabs}
\usepackage{multirow}
\usepackage[table]{xcolor}
\usepackage{array}
\usepackage{makecell}
\usepackage{pifont}
\usepackage{enumitem}

\usepackage{cuted}     
\usepackage{capt-of}   

\def\wacvPaperID{1952} 
\def\confName{WACV}
\def\confYear{2026}

\title{ArchitectHead: Continuous Level of Detail Control for 3D Gaussian Head Avatars}

\author{
Peizhi Yan$^{1}$,
Rabab Ward$^{1}$,
Qiang Tang,
Shan Du$^{3}$\thanks{Corresponding author: Shan Du.}\\
$^{1}$University of British Columbia {\tt\small \{yanpz, rababw\}@ece.ubc.ca}\\
$^{3}$University of British Columbia (Okanagan) {\tt\small shan.du@ubc.ca}
}

\begin{document}
\maketitle

\begin{strip}
\centering
\includegraphics[width=\textwidth]{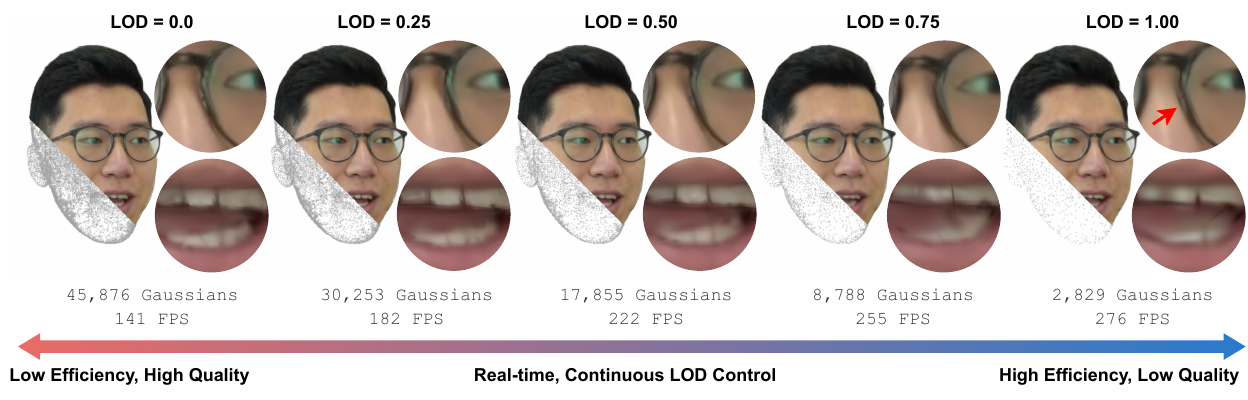}
\vspace{-0.65cm} 
\captionof{figure}{ArchitectHead supports continuous LOD control ranging from 0 (highest) to 1.0 (lowest). This figure shows the rendered image quality under different LOD settings. For each case, we provide a zoom-in view of selected regions. We recommend viewing the figure on a digital device with zoom for better inspection. The grey dots indicate the positions of the 3D Gaussian points. The test identity is taken from the NeRFace dataset \cite{gafni2021dynamic}. Red arrow indicates visible artifacts (sparse Gaussians) in the lowest LOD.}
\label{fig:teaser}
\end{strip}


\input{sec/0_abstract}

\input{sec/1_intro}

\input{sec/2_related}

\input{sec/3_method}

\input{sec/4_experiment}

\input{sec/5_conclusion}

{
    \small
    \bibliographystyle{ieeenat_fullname}
    \bibliography{main}
}

\end{document}

%% file: sec/0_abstract.tex
\begin{abstract}
3D Gaussian Splatting (3DGS) has enabled photorealistic and real-time rendering of 3D head avatars. Existing 3DGS-based avatars typically rely on tens of thousands of 3D Gaussian points (Gaussians), with the number of Gaussians fixed after training. However, many practical applications require adjustable levels of detail (LOD) to balance rendering efficiency and visual quality. In this work, we propose ``ArchitectHead", \textbf{the first framework} for creating 3D Gaussian head avatars that support continuous control over LOD. Our key idea is to parameterize the Gaussians in a 2D UV feature space and propose a UV feature field composed of multi-level learnable feature maps to encode their latent features. A lightweight neural network-based decoder then transforms these latent features into 3D Gaussian attributes for rendering. ArchitectHead controls the number of Gaussians by dynamically resampling feature maps from the UV feature field at the desired resolutions. This method enables efficient and continuous control of LOD without retraining. Experimental results show that ArchitectHead achieves state-of-the-art (SOTA) quality in self and cross-identity reenactment tasks at the highest LOD, while maintaining near SOTA performance at lower LODs. At the lowest LOD, our method uses only 6.2\% of the Gaussians while the quality degrades moderately (L1 Loss +7.9\%, PSNR –0.97\%, SSIM –0.6\%, LPIPS Loss +24.1\%), and the rendering speed nearly doubles. The code will be released. Project homepage: \url{https://peizhiyan.github.io/docs/architect/}.
\end{abstract}

%% file: sec/1_intro.tex
\vspace{-0.5cm}

\section{Introduction}
\label{sec:intro}

3D head avatars are an important component of digital humans, with applications ranging from VR, telepresence, video gaming, to video conferencing. While recent advancements in 3D Gaussian Splatting (3DGS) \cite{kerbl20233d} have achieved a higher degree of photorealism than traditional mesh-based methods, they are more computationally intensive and struggle to maintain real-time performance when multiple avatars are rendered simultaneously \cite{qian2024gaussianavatars, xiang2023flashavatar, shao2024splattingavatar, zhang2025hravatar, li2025rgbavatar}. In computer graphics, level of detail (LOD) techniques for 3D meshes are widely used to balance visual quality and efficiency \cite{hoppe2023progressive}. However, applying LOD to 3DGS is challenging due to the unstructured nature of Gaussian point clouds and the need for dynamic animation in head avatars. To address this gap, we propose ``ArchitectHead", a framework that learns 3D Gaussian head avatars from monocular videos and allows LOD adjustment after training. Unlike conventional LOD methods that provide only a few discrete levels (typically 3 to 4 levels) \cite{seo2024flod, dongye2024lodavatar}, ArchitectHead supports continuous LOD control, making it easier to balance rendering speed and visual quality.


LOD methods are used to simplify 3D object geometries to improve rendering efficiency and save computational resources. For example, distant or less important parts of a 3D scene can be rendered with reduced geometric detail. In general, there are three types of LOD methods: (1) discrete, (2) continuous, and (3) view-dependent \cite{heok2004review}. The discrete LOD method creates a small set of simplified versions of the same 3D model offline, which is easy to implement but often causes unsmooth visual effects when switching between levels \cite{seo2024flod, dongye2024lodavatar}. Continuous LOD methods allow progressive adjustment of LOD at run time, which leads to smoother transitions and better use of resources \cite{hoppe2023progressive, takikawa2021neural}. View-dependent LOD further adapts the level of detail based on the viewing position, allocating higher detail to visible or important regions of a model \cite{schutz2019real}. In this work, our focus is to bring the advantages of continuous LOD into 3D Gaussian head avatars, where smooth and flexible control over level of detail is essential.


3D Gaussian head avatars represent the head using Gaussian points, which are rendered with the 3DGS rasterization method \cite{kerbl20233d} to produce photorealistic images. Since head avatars must be animatable, existing methods typically bind the 3D Gaussians to a mesh-based head model to provide controllability. Two major binding strategies have been explored. The first associates Gaussians with the triangle facets of the mesh, so that transformations applied to the facets are directly propagated to the corresponding Gaussians \cite{qian2024gaussianavatars, li2025rgbavatar, tang2025gaf, zheng2025headgap}. The second uses a 2D UV map as an intermediate representation to connect the mesh geometry with the Gaussians. In this case, the mesh geometry is first rasterized onto a UV map with predefined 2D-3D correspondences, where each pixel encodes the 3D location of the mesh surface, and these locations are then used to initialize the Gaussian positions \cite{xiang2023flashavatar, li2024uravatar, yan2025gaussian, zhang2025fate, song2025streamme}. We follow the second strategy in designing ArchitectHead, as the UV map provides an inherent 2D representation in which nearby Gaussian points are also neighbors in UV space. This property makes it natural to control the number of Gaussians by adjusting the resolution of the UV map.

Although the UV-based strategy provides a natural way to control the number of Gaussians through the UV map resolution, there are two remaining challenges. First, the UV position map alone does not capture sufficient local information to represent detailed 3D head appearance. Second, it is necessary to balance different resolutions while maintaining smooth transitions across LODs. These challenges motivate our design of ``ArchitectHead" in this work.

ArchitectHead is the first head avatar creation framework that supports continuous adjustment of the level of detail. We formulate continuous LOD control as resampling from the 2D UV feature space. The key idea is to make controlling the number of Gaussians in a head avatar as simple as resizing a 2D image. To achieve this, we design a learnable multi-level UV feature field, structured as a pyramid of UV-anchored feature maps at different resolutions. The feature maps store the per-Gaussian latent features, which will be learned during training. Our model is trained per-person following the training objective of existing monocular head avatar works to gain better personalization \cite{xiang2023flashavatar, li2025rgbavatar, dhamo2024headgas, chen2023monogaussianavatar}. After training, we can sample a UV feature map of any resolution between the maximum and minimum levels to achieve run-time LOD control. Our sampling strategy assigns higher weights to feature maps with resolutions closer to the desired one to blend the UV feature maps of different resolutions. This weighted blending method ensures smooth transitions across LODs. The sampled features are concatenated with positionally encoded point locations, the expression code, and the LOD value to form the latent representation of the Gaussian points. We employ a lightweight neural decoder to convert these latent features into 3D Gaussian attributes, ready for rendering. 

The training has two stages. In the first stage, we train with the highest LOD (LOD 0) to capture fine details and stabilize learning. In the second stage, we randomly sample LOD values between 0.0 and 1.0 so the decoder can adapt to different levels of detail. This design enables efficient and flexible LOD control for 3D Gaussian head avatars without the need for retraining. Figure~\ref{fig:teaser} shows an example of our reconstructed head avatar with different LOD settings.

In summary, our main contributions are:
\begin{enumerate}
    \item We propose ArchitectHead, a novel framework for creating 3D Gaussian head avatars that support on-the-fly control of LOD without re-training. To our knowledge, this is the first 3D Gaussian head avatar method that enables real-time continuous LOD adjustment, providing greater scalability for balancing rendering quality and efficiency across applications.
    \item We introduce a learnable multi-level UV feature field along with an adaptive sampling strategy to capture spatial information across high to low resolutions. This ensures smooth transitions between different LODs.
    \item We conduct extensive experiments on monocular video datasets for 3D head avatar reconstruction, demonstrating that ArchitectHead achieves effective and scalable control of LOD while maintaining photorealistic quality.
\end{enumerate}

%% file: sec/2_related.tex
\section{Related Works}
\label{sec:related}

\subsection{3D Head Avatars}

Existing 3D head avatar methods can be categorized by their 3D representation. Mesh-based methods are the most traditional, known for their rendering efficiency \cite{hu2017avatar, ma2021pixel, egger20203d}. A notable example is the 3D Morphable Models (3DMMs), which are created by registering real 3D face scans to a template mesh \cite{blanz2023morphable, li2017learning, egger20203d}. While 3DMMs provide disentangled identity and expression representations, they are less effective at modeling non-rigid facial features like hair. Neural radiance field (NeRF)-based methods \cite{mildenhall2020nerf} use neural networks to learn latent 3D representations and render images by sampling points in 3D space, estimating transmittance, and accumulating through volume rendering \cite{teotiahq3davatar, sun2023next3d, wang2024lightavatar}. These methods support learning from images and videos and can achieve photorealistic results. However, they are computationally intensive due to the need for extensive point sampling per frame and are less accurate with geometry. Point-based methods explicitly represent the head with 3D points, allowing faster rendering and more flexibility than mesh-based approaches \cite{zheng2023pointavatar, chen2023monogaussianavatar, qian2024gaussianavatars, li2025rgbavatar}. Among the point-based methods, 3D Gaussian splatting-based methods have recently become the most popular, which we will detail in the following subsection.

\subsection{3D Gaussian-based Head Avatars}

3D Gaussian Splatting (3DGS) \cite{kerbl20233d} has emerged as an efficient and photorealistic representation for 3D head avatars, capable of real-time rendering. To enable reconstruction and animation, many works bind Gaussians to a 3DMM-based prior such as FLAME \cite{li2017learning}. This can be done by rigging them to mesh triangles \cite{qian2024gaussianavatars, li2025rgbavatar, zheng2025headgap} or initializing them via UV maps \cite{xiang2023flashavatar, li2024uravatar, song2025streamme, yan2025gaussian}. Several methods further improve expressiveness through neural decoders or expression blendshapes, enabling more controllable facial dynamics \cite{wang2025gaussianhead, dhamo2024headgas, ma20243d, yan2025gaussian, li2025rgbavatar}. Another research direction focuses on generalization, using large-scale prior models for few-shot or single-image personalization \cite{li2024uravatar, he2025lam, kirschstein2025avat3r, zheng2025headgap}. Diffusion-based frameworks have also been explored to improve monocular video supervision and robustness to novel views \cite{tang2025gaf, taubner2025cap4d, gerogiannis2025arc2avatar}. Despite these advancements, a common limitation persists, as all these methods use a fixed number of Gaussians after training, making Level of Detail (LOD) control challenging. Our work is the first to introduce continuous LOD control for 3DGS head avatars, allowing for a dynamic balance between rendering efficiency and visual quality to suit various practical needs. 

UV-based FlashAvatar \cite{xiang2023flashavatar} is similar to our method. It uses a multi-layer perceptron (MLP) decoder to generate expression-conditioned 3D Gaussians by taking a positionally encoded UV mesh geometry map and expression coefficients from FLAME \cite{li2017learning} as input. Despite its simple yet effective solution, it does not support changing the UV map resolution after training. This is because 3D Gaussians hold complex information, including unique rotations and sizes, which prevents them from being easily scaled like images. Our work addresses this by proposing a learnable and scalable UV-based feature field that enables the decoder to adapt to varying UV resolutions.

\subsection{Level of Detail in 3D Gaussian Splatting}

Recent research has extended level of detail concepts to 3DGS, but these methods are mainly for static scenes such as urban environments \cite{liu2024citygaussian, windisch2025lod}. Some methods introduce the hierarchical structures, enabling efficient rendering by pruning distant or less important Gaussians \cite{ren2025octree, kulhanek2025lodge, shen2025lod}. Other methods, such as CityGaussian and FLoD, focus on scalable multi-level representations to adapt to hardware constraints \cite{liu2024citygaussian, seo2024flod}. LOD-GS applies the filtering mechanism that is sensitive to sampling rate to improve visual quality across different zoom levels \cite{yang2025lod}. Milef et al. propose a continuous LOD method that learns to rank splats by importance to enable efficient distance-based rendering without inference-time overhead \cite{milef2025learning}. Although these methods are effective for large static scenes, they are less suitable for 3D head avatars, which are highly dynamic and require controllability to support facial animation. 

The most relevant method to ours is LoDAvatar, which adapts mesh triangle-bound Gaussians for full-body avatars using hierarchical embedding and selective detail enhancement \cite{dongye2024lodavatar}. However, LoDAvatar only supports discrete LOD control and relies on synthetic multi-view images for training, making it less suitable for personalized avatar creation from monocular video. In contrast, our work supports continuous LOD control for 3D Gaussian head avatars, making the change of LOD smooth while also allowing the training from monocular videos for better photorealism.

%% file: sec/3_method.tex
\section{Method}
\label{sec:method}

\begin{figure*}[ht]
    \centering
    \includegraphics[width=1.0\linewidth]{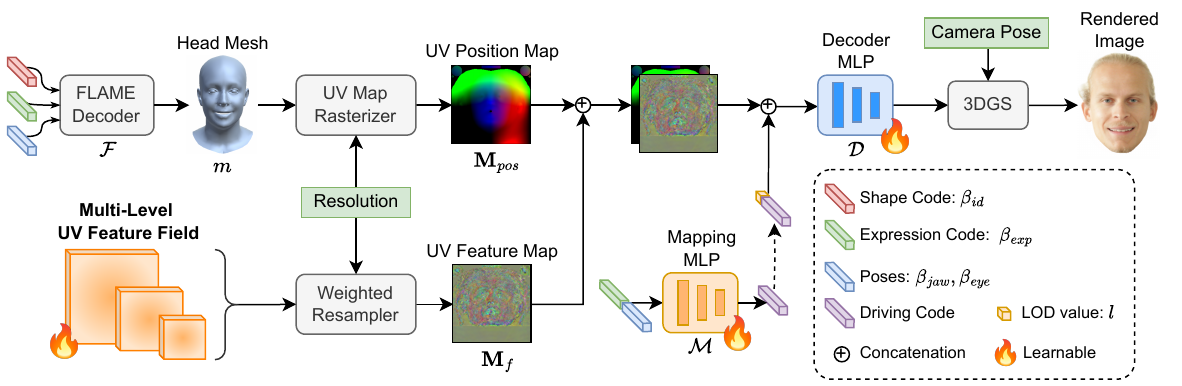}
    \vspace{-0.5cm}
    \caption{\textbf{Pipeline of ArchitectHead.} We propose a 3D Gaussian head avatar creation method with continuous level of detail (LOD) control. Starting from shape and expression codes, we use the FLAME head model to generate the 3D mesh geometry, which is rasterized into a UV position map at the desired resolution. A multi-level UV feature field is introduced to learn local latent features, from which our weighted resampler extracts a UV feature map of the target resolution. This map is concatenated with the UV position map, the desired LOD value, and a driving code obtained from expression and pose codes via an MLP network $\mathcal{M}$. The resulting pixel-wise latent features are decoded by an MLP-based decoder into 3D Gaussian attributes, which are rendered using 3D Gaussian Splatting (3DGS).}
    \label{fig:framework}
\end{figure*}

\subsection{Preliminaries: 3D Gaussian Splatting}

3D Gaussian Splatting (3DGS) represents a 3D scene using anisotropic 3D Gaussian primitives and renders them through a splatting-based process that is both differentiable and efficient \cite{kerbl20233d}. The influence of a Gaussian at a 3D position $\mathbf{x} \in \mathbb{R}^3$ is defined by a modified Gaussian probability density function:

\begin{equation}
    G(\mathbf{x}) = e^{-\frac{1}{2} (\mathbf{x} - \mathbf{\mu})^{\top} \mathbf{\Sigma}^{-1} (\mathbf{x} - \mathbf{\mu})},
    \label{eq:3dgaussian}
\end{equation}
where $\mathbf{\mu}$ is the Gaussian center and $\mathbf{\Sigma} \in \mathbb{R}^{3 \times 3}$ its covariance. To guarantee $\mathbf{\Sigma}$ is positive semi-definite, it is constructed from a rotation matrix $\mathbf{R}$ and a scaling matrix $\mathbf{S}$:

\begin{equation}
    \mathbf{\Sigma} = \mathbf{R} \mathbf{S} \mathbf{S}^{\top} \mathbf{R}^{\top}.
    \label{eq:covariance}
\end{equation}

This corresponds to an oriented ellipsoid, parameterized by a scale vector $\mathbf{s}$ and a quaternion $\mathbf{q}$ for 3D rotation to avoid Gimbal lock. Each Gaussian is thus defined as $\mathbf{g} = \{\mathbf{\mu} \in \mathbb{R}^{3}, \mathbf{s} \in \mathbb{R}^{3}, \mathbf{q} \in \mathbb{R}^{4}, \alpha \in \mathbb{R}^{1}, \mathbf{c} \in \mathbb{R}^{\mathbf{SH}} \}$. $\alpha$ is its opacity and $\mathbf{c}$ is the Spherical Harmonic (SH) coefficients for computing the color \cite{fridovich2022plenoxels}. During rendering, the Gaussians are projected to the 2D image plane and composited using alpha blending, similar to volume rendering \cite{mildenhall2020nerf}.

\subsection{Framework Overview}

As shown in Figure~\ref{fig:framework}, we leverage the geometry prior of the FLAME head model \cite{li2017learning} to initialize the positions of 3D Gaussians. The 3D head mesh $m$ is derived as $m = \mathcal{F}(\beta_{id}, \beta_{exp}, \beta_{jaw}, \beta_{eye})$, where $\mathcal{F}$ is the FLAME decoder, $\beta_{id} \in \mathbb{R}^{300}$, $\beta_{exp} \in \mathbb{R}^{100}$, $\beta_{jaw} \in \mathbb{R}^{3}$, and $\beta_{eye} \in \mathbb{R}^{6}$ are the identity code, expression code, jaw pose, and eye poses, respectively. Following \cite{qian2024gaussianavatars}, we add teeth to the original FLAME mesh to better initialize Gaussians in the mouth region. The mesh geometry is rasterized into a UV map. We then apply positional encoding following \cite{xiang2023flashavatar} to obtain $\mathbf{M}_{pos} \in \mathbb{R}^{S \times S \times D_{\textit{PE}}}$, where $S$ is the desired UV resolution and $D_{\textit{PE}}$ is the dimensionality after positional encoding.

Denote the continuous LOD level as $0 \leq l \leq 1.0$, where $l = 0$ corresponds to the highest LOD and $l = 1.0$ to the lowest. Also, define $S_{max}$ and $S_{min}$ as the maximum and minimum UV map resolutions, respectively. The desired UV map resolution is then computed as:
\begin{equation}
    S = S_{\max} - l \,(S_{\max} - S_{\min}).
    \label{eq:resolution}
\end{equation}

The UV feature map $\mathbf{M}_{f}  \in \mathbb{R}^{S \times S \times D_{f}}$ , where $D_{f}$ is the feature dimension, is resampled from the proposed multi-level UV feature field (detailed in Section~\ref{sec:multi-level-uv-field}). This map is then concatenated channel-wise with a positionally encoded UV position map, $\mathbf{M}_{pos}$, to inject various frequency-based information into the features. This approach, leveraging the learnable feature map $\mathbf{M}_{f}$, significantly improves the model's ability to represent local details, which in turn leads to enhanced visual quality.  

Inspired by \cite{li2025rgbavatar}, we use a small multi-layer perceptron (MLP) network $\mathcal{M}$ to learn the mapping from higher dimension codes ($\beta_{exp}$, $\beta_{jaw}$, and $\beta_{eye}$) to a lower dimension driving code with $K$ dimensions. This code conditions the decoder network on facial expressions. The mapping helps the model focus on the most important expression variations and reduces overfitting to specific expression or pose codes. We then concatenate the driving code with the LOD value $l$, creating a one-dimensional vector. This vector is then appended to each pixel of the UV map, resulting in a final feature map with $D_{\textit{PE}} + D_{f} + K + 1$ channels.

The decoder network consists of five MLPs, each of which is used for decoding the per-Gaussian UV feature into one of the 3D Gaussian attributes: ${\Delta\mathbf{\mu}, \mathbf{s}, \mathbf{q}, \alpha, \mathbf{c}}$. The final Gaussian location is computed as $\mathbf{\mu} = \Delta\mathbf{\mu} + \mathbf{\mu}_{\textit{init}}$, where $\mathbf{\mu}_{\textit{init}}$ is the initial position from the UV position map. Given the camera pose, the decoded 3D Gaussians are rendered through the 3DGS rasterizer to derive the image. Please refer to the supplementary material for details.

\subsection{Multi-Level UV Feature Field}
\label{sec:multi-level-uv-field}

Existing works have shown that using higher-dimensional learnable 3D Gaussian features can produce better visual quality \cite{hu2025tgavatar, serifi2025hypergaussians, giebenhain2024npga}. This is because higher-dimensional features provide each 3D Gaussian point with greater representational capacity. Motivated by this, we propose using a 2D UV feature map to encode and learn these high-dimensional, per-Gaussian features. However, this approach presents a challenge when implementing continuously controllable LOD if the UV feature map is simply resized. Since each pixel in the map corresponds to a 3D Gaussian point, neighboring Gaussians can have significantly different rotations and scales. While downsampling from a high to a low resolution can merge features of nearby Gaussians, it also smooths out critical information, causing adjacent Gaussians to share overly similar attributes and compromising local detail.

To address the above-mentioned issue, we propose to use multiple UV feature maps of different resolutions to form a multi-level feature field. We can resample a UV feature map of any size between the largest and smallest UV feature maps in the multi-level feature field. Define the feature field as a set of UV feature maps $ \mathcal{U} = \{ \mathbf{M}_{f}^{S_{1}}, ... , \mathbf{M}_{f}^{S_{N}}\}$ ordered by the resolutions $S_{1}, ..., S_{N}$ increasingly. $\mathcal{U}$ contains at least the feature maps of the maximum and minimum resolutions. To sample the feature map of a given resolution $S$ ($S_{min} \leq S \leq S_{max}$), we first resize all the feature maps to the resolution of $S$ using bi-cubic interpolation. Then we blend all the resized feature maps via the following formula:
\begin{equation}
    \mathbf{M}_{f} = \sum_{i=1}^{N} w_{i} * \mathcal{I}(\mathbf{M}_{f}^{S_{i}}, S), 
    \label{eq:blend}
\end{equation}
where $w_i$ is the blending weight for the $i^{\textit{th}}$ feature map $\mathbf{M}_{f}^{S_{i}}$, and $\mathcal{I}(\mathbf{M}_{f}^{S_{i}}, S)$ is the bi-cubic interpolation function that resizes $\mathbf{M}_{f}^{S_{i}}$ to a given resolution $S$. 

The computation of $w_i$ is defined as follows:
\begin{equation}
    w_{i} \;=\;
    \frac{
        \exp\!\left(-|r_i - r|/\tau\right)
    }{
        \sum_{j=1}^{N}
        \exp\!\left(-|r_j - r|/\tau\right)
    },
    \label{eq:blend_weight}
\end{equation}
where $r_i = log_{e}S_i$, $r = log_{e}S$, $\tau > 0$ is the temperature controlling the softness of interpolation. Equation~\ref{eq:blend_weight} is a softmax-based formulation, which gives higher weight to the feature map whose resolution is closer to $S$, while ensuring $w_i$ never reaches zero so that gradients can always propagate properly.

\subsection{Loss Functions}

The training loss is defined as:
\begin{equation}
    \mathcal{L} = \mathcal{L}_{rgb} + \lambda_{lpips}\mathcal{L}_{lpips} + \lambda_{\mu}\mathcal{L}_{\mathbf{\mu}} + \lambda_{\mathbf{s}}(1 -  0.5l)\mathcal{L}_{\mathbf{s}},
    \label{eq:loss}
\end{equation}
where $\mathcal{L}_{rgb}$ is the image reconstruction loss, $\mathcal{L}_{lpips}$ is the LPIPS perceptual loss \cite{zhang2018unreasonable, xiang2023flashavatar}, $\mathcal{L}_{\mathbf{\mu}}$ and $\mathcal{L}_{\mathbf{s}}$ are regularization terms on the Gaussian point location offsets and scales, respectively, and $\lambda$s are their weights. Both $\mathcal{L}_{\mathbf{\mu}}$ and $\mathcal{L}_{\mathbf{s}}$ are computed using the mean L2 norm of the location offsets and scales. For $\mathcal{L}_{\mathbf{s}}$, we assign more weights to smaller $l$ (higher LOD) because the Gaussians are denser. Specifically, $\mathcal{L}_{rgb}$ has two sub-terms:
\begin{equation}
    \mathcal{L}_{rgb} = \mathcal{L}_{H}(I, \hat{I}) +  \lambda_{parts}\mathcal{L}_{H}(I \cdot \mathbf{B}, \hat{I} \cdot \mathbf{B}),
    \label{eq:rgb}
\end{equation}
where $\mathcal{L}_{H}$ is the Huber loss \cite{huber1992robust}, $I$ is the ground-truth image, $\hat{I}$ is the rendered image, and $\mathbf{B}$ is the binary mask of eyes and mouth parts.

\subsection{Training Scheme}

The training has two stages. In the first stage, we use the highest Level of Detail (LOD) to jointly optimize the multi-level latent field along with the mapping network, $\mathcal{M}$ and the decoder network $\mathcal{D}$. The second stage of training focuses on the model's ability to handle continuous detail. In this stage, we randomly sample LODs from the range $[0, 1.0]$ and accumulate their loss gradients for optimization, while keeping the mapping network $\mathcal{M}$ fixed.

%% file: sec/4_experiment.tex
\section{Experiments}
\label{sec:experiment}

\begin{figure*}[ht]
    \centering
    \includegraphics[width=1.0\linewidth]{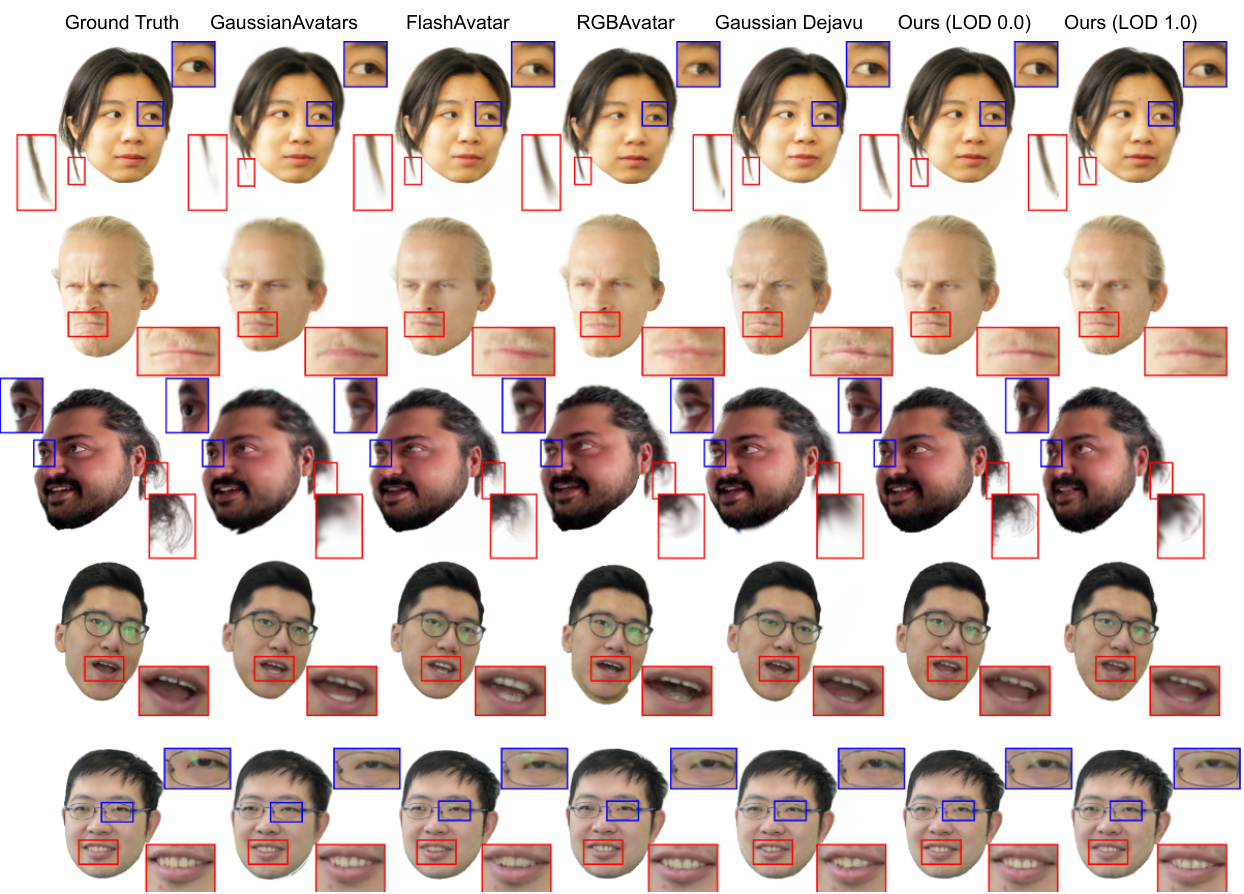}
    \caption{\textbf{Qualitative comparisons of self-reenactment results.} Selected regions are zoomed in for clearer comparison of fine details. The last two columns show our method with the highest (LOD=0.0) and lowest (LOD=1.0) settings. Compared to existing methods, our approach preserves finer details at the highest LOD while also maintaining reasonable quality at the lowest LOD.}
    \label{fig:quality_self}
\end{figure*}

\subsection{Datasets}

We use two monocular video datasets in our experiments: the PointAvatar dataset \cite{zheng2023pointavatar} and the INSTA dataset \cite{zielonka2023instant}. The PointAvatar dataset contains videos of three subjects, while the INSTA dataset provides ten videos of different subjects. Since one subject in INSTA overlaps with PointAvatar, we use only nine videos from INSTA. For each video, the first 90\% of frames are used for training and the remaining 10\% for evaluation.

\subsection{Implementation Details}

We implement our method using \textit{PyTorch}. To render the 3D Gaussians, we use the open-source 3DGS rasterizer \textit{gsplat} \cite{ye2025gsplat}. The maximum and minimum UV map resolutions are set to $S_{max} = 256$ and $S_{max} = 64$, respectively. We use 12 frequencies for both sine and cosine in the positional encoding, resulting in $D_{pos} = 75$ dimensions (see supplementary materials). Following \cite{li2025rgbavatar}, we set the reduced driving code dimension to $K=20$. The learnable feature maps have $D_{f} = 64$ channels. The multi-level feature field consists of three resolutions, $256\times 256$, $128\times 128$, and $64\times 64$, respectively. The temperature in Equation~\ref{eq:blend_weight} is set to $\tau = 0.35$. We use $\mathbf{SH} = 3$ in our work. We implement a fitting-based FLAME head tracker to derive the FLAME codes and camera pose for each frame in the monocular videos. 

\textbf{Training:} Training is conducted on a single Nvidia RTX 4090 GPU with a batch size of one. In stage one, we train for 15,000 steps. In stage two, we render five LOD levels at each step and train for 30,000 steps in total. The training time is approximately 30 minutes for stage one and 2.5 hours for stage two. Please refer to supplementary materials for details. The loss term weights are $\lambda_{\textit{parts}} = 20$, $\lambda_{\textit{lpips}} = 0.05$, $\lambda_{\mathbf{\mu}} = 0.001$, and $\lambda_{\mathbf{s}} = 0.5$.

\subsection{Baselines}

The baselines are state-of-the-art Gaussian head avatar methods \cite{qian2024gaussianavatars, xiang2023flashavatar, li2025rgbavatar, yan2025gaussian}. Both GaussianAvatars \cite{qian2024gaussianavatars} and RGBAvatar \cite{li2025rgbavatar} bind Gaussians to triangle facets, while FlashAvatar \cite{xiang2023flashavatar} and Gaussian Dejavu \cite{yan2025gaussian} adopt UV-based representations. In addition, both RGBAvatar and Gaussian Dejavu employ Gaussian blendshapes to achieve expressive animation. These methods are representative of the major categories of Gaussian head avatars, including mesh-bound \cite{qian2024gaussianavatars, li2025rgbavatar}, UV-based \cite{xiang2023flashavatar, yan2025gaussian}, and blendshape-based \cite{yan2025gaussian, li2025rgbavatar}.

\subsection{Self and Cross-Identity Reenactment}

\begin{figure}[ht]
    \centering
    \includegraphics[width=1.0\linewidth]{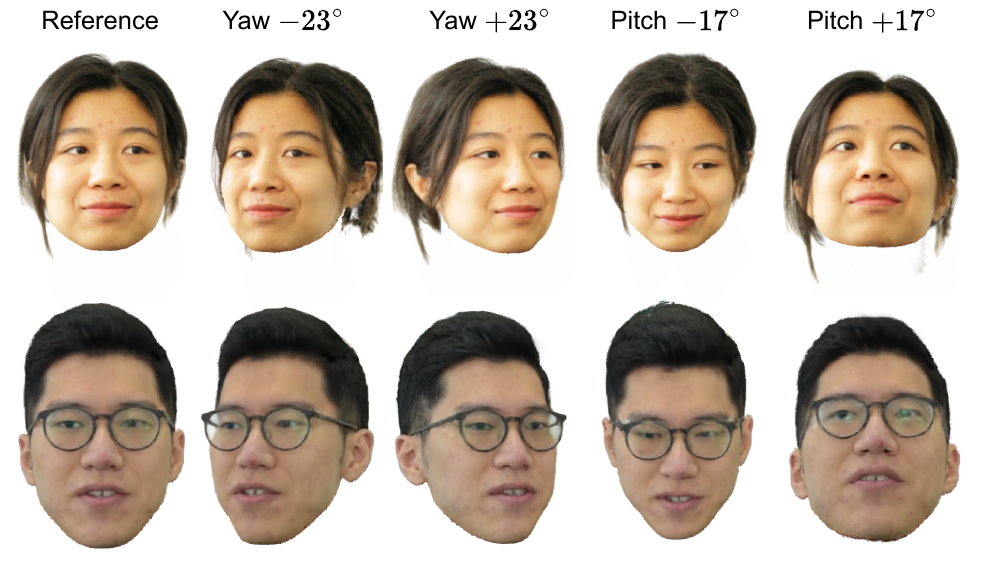}
    \caption{\textbf{Rendered novel views.} The reference images (left-most) are rendered with the default camera pose, while the other images are rendered with yaw or pitch angle offsets using an orbit camera.}
    \label{fig:novel_views}
\end{figure}

\textbf{Self-reenactment.} The self-reenactment task involves using FLAME codes from a specific identity to drive an avatar model that was trained on that same identity's video. We conduct experiments on both the PointAvatar and INSTA datasets. The quantitative results in Tables~\ref{tab:self_imavatar} and \ref{tab:self_insta} show that the proposed method achieves the best performance on both datasets across all metrics, including L1 distance, PSNR, SSIM, and perceptual loss (LPIPS). We also evaluated the trained avatar in half-precision (fp16), and the results are comparable to those obtained with full precision. Figure~\ref{fig:quality_self} presents qualitative results. At the highest LOD setting, our method captures finer details, while at the lowest LOD setting it can still represent reasonable head shapes with sparse Gaussian points. In addition, Figure~\ref{fig:novel_views} illustrates novel views rendered from the reconstructed avatars.

\begin{table}[h]
\centering
\small
\begin{tabular}{l@{\hskip 8pt}c@{\hskip 6.5pt}c@{\hskip 6.5pt}c@{\hskip 6.5pt}c}
\hline
\textbf{Method} & \textbf{L1} $\downarrow$ & \textbf{PSNR} $\uparrow$ & \textbf{SSIM} $\uparrow$ & \textbf{LPIPS} $\downarrow$ \\
\hline
GaussianAvatars \cite{qian2024gaussianavatars}       & 0.048 & 26.009 & 0.916 & 0.112 \\
FlashAvatar \cite{xiang2023flashavatar}              & 0.044 & 27.223 & 0.915 & 0.070 \\
RGBAvatar \cite{li2025rgbavatar}                     & 0.053 & 28.106 & 0.907 & 0.100 \\
Gaussian Dejavu \cite{yan2025gaussian}               & 0.045 & 27.819 & 0.923 & 0.065 \\
Ours (fp16)    & \textbf{0.038} & 28.615 & \textbf{0.928} & \textbf{0.058} \\
Ours (best)    & \textbf{0.038} & \textbf{28.621} & \textbf{0.928} & \textbf{0.058} \\
\hline
\end{tabular}
\caption{\textbf{Quantitative comparisons of self-reenactment results} on PointAvatar dataset.}
\label{tab:self_imavatar}
\end{table}

\begin{table}[h]
\centering
\small
\begin{tabular}{l@{\hskip 8pt}c@{\hskip 6.5pt}c@{\hskip 6.5pt}c@{\hskip 6.5pt}c}
\hline
\textbf{Method} & \textbf{L1} $\downarrow$ & \textbf{PSNR} $\uparrow$ & \textbf{SSIM} $\uparrow$ & \textbf{LPIPS} $\downarrow$ \\
\hline

GaussianAvatars \cite{qian2024gaussianavatars}      & 0.030 & 26.979 & 0.945 & 0.069 \\
FlashAvatar \cite{xiang2023flashavatar}             & 0.027 & 28.362 & 0.947 & 0.039 \\
RGBAvatar \cite{li2025rgbavatar}                    & 0.037 & 27.583 & 0.924 & 0.060 \\
Gaussian Dejavu \cite{yan2025gaussian}              & 0.024 & 29.722 & 0.956 & 0.034 \\
Ours (fp16)     & \textbf{0.022} & 30.379 & \textbf{0.960} & \textbf{0.032} \\
Ours (best)     & \textbf{0.022} & \textbf{30.389} & \textbf{0.960} & \textbf{0.032} \\
\hline
\end{tabular}
\caption{\textbf{Quantitative comparisons of self-reenactment results} on INSTA dataset.}
\label{tab:self_insta}
\end{table}

\begin{figure}[ht]
    \centering
    \includegraphics[width=1.0\linewidth]{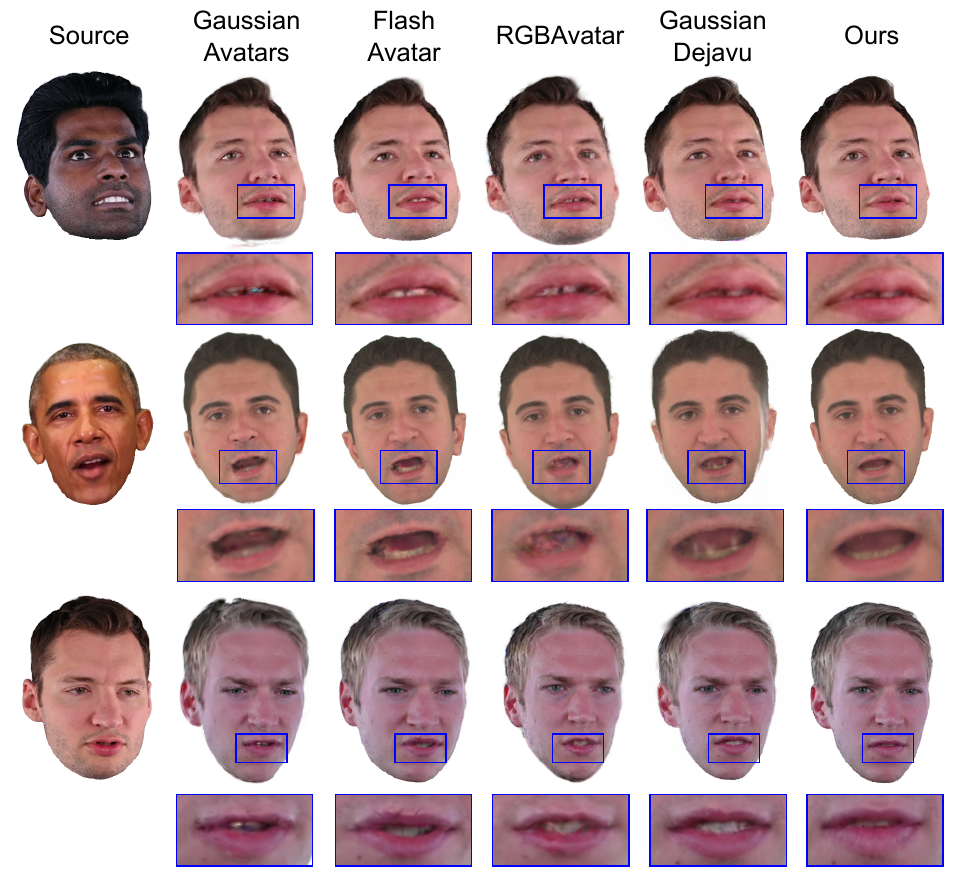}
    \caption{\textbf{Qualitative results of cross-identity reenactment.} The first column shows the source images that provide the expression codes and camera poses. The remaining columns in each row present the trained head avatar of another individual reenacted using the source codes and poses.}
    \label{fig:quality_cross}
\end{figure}

\noindent\textbf{Cross-Identity Reenactment.} In cross-identity reenactment, we use FLAME codes from the video of one identity to drive avatars trained on another identity. Figure~\ref{fig:quality_cross} shows qualitative results. Our method achieves higher fidelity and better visual quality compared to the baselines.

\subsection{Level of Detail Control}

\begin{figure}[ht]
    \centering
    \includegraphics[width=1.0\linewidth]{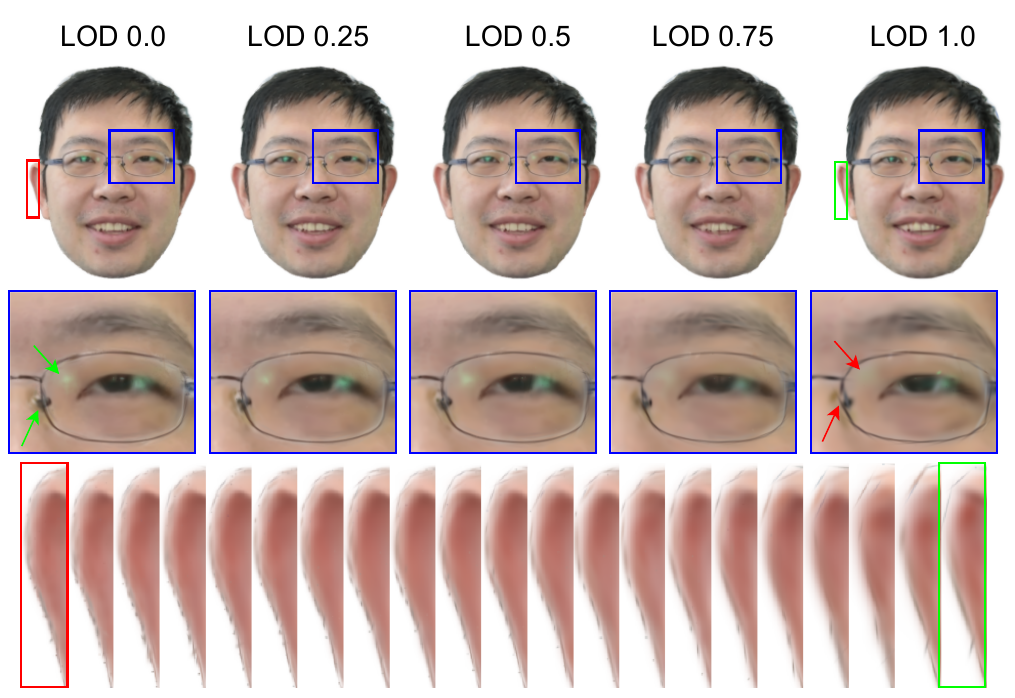}
    \caption{\textbf{Qualitative results of varying LODs.} The second row shows zoomed-in regions highlighted by the blue rectangle in the first row. \textcolor{red}{Red} arrows indicate sub-optimal results (missing glare and fine details), compared to the corresponding locations marked by \textcolor{green}{green} arrows in our highest LOD (LOD=0) result. The third row illustrates the same region (ear) rendered with different LODs from 0.0 to 1.0 in steps of 0.05. }
    \label{fig:varying_lods}
\end{figure}

In this experiment, we demonstrate the LOD control capability of the proposed method. Figure~\ref{fig:varying_lods} shows the trained avatar model rendered under different LOD settings. We observe that while the overall differences are subtle, the quality decreases in zoomed-in regions at lower LODs. Therefore, for scenes where the camera is close to the avatar, we recommend using a higher LOD to achieve better visual quality. Conversely, if the camera is far away or quality is less critical, a lower LOD can be used to save computational resources. We also benchmark the rendering speed across different LOD levels, with results summarized in Table~\ref{tab:fps}. Using half-precision (fp16) leads to significantly faster rendering at higher LODs. Importantly, the rendering quality does not degrade noticeably with fp16, even though the model is trained in full precision (fp32); detailed comparisons are provided in the supplementary materials.

In Table~\ref{tab:ablation}, we evaluate different LOD settings. At the lowest LOD ($l=1.0$), the model uses only 6.2\% of the Gaussians compared to the highest LOD, with a small increase in L1 loss (+7.9\%), and minor drops in PSNR (–0.97\%) and SSIM (–0.6\%), while nearly doubling rendering speed (+96\% FPS).

\begin{table}[h]
\centering
\small
\begin{tabular}{c|r|c|c|c|c}
\hline
\textbf{LOD} & \textbf{\#Gaussians} & \makecell{\textbf{4090}\\\textbf{(fp16)}} & \makecell{\textbf{4090}\\\textbf{(fp32)}} & \makecell{\textbf{A6000}\\\textbf{(fp16)}} & \makecell{\textbf{A6000}\\\textbf{(fp32)}} \\
\hline
0.00  & 45,876 & 141 &  57 &  64 &  38 \\
0.25  & 30,253 & 182 & 132 &  86 &  61 \\
0.50  & 17,855 & 222 & 181 & 113 &  87 \\
0.75  &  8,788 & 255 & 245 & 146 & 123 \\
1.00  &  2,829 & 276 & 301 & 142 & 160 \\
\hline
\end{tabular}
\caption{Rendering FPS $\uparrow$ (frames per second) performance comparison across LOD settings, GPUs, and floating-point precisions. \textit{\#Gaussians} denotes the number of 3D Gaussian points.}
\label{tab:fps}
\end{table}

\subsection{Ablation Studies}

\begin{table}[h]
\definecolor{third}{RGB}{255,255,255}   
\definecolor{second}{RGB}{211,255,255}  
\definecolor{best}{RGB}{60,244,255}     
\newcommand{\A}{\cellcolor{best}{}}
\newcommand{\B}{\cellcolor{second}{}}
\newcommand{\C}{\cellcolor{third}{}} 
\DeclareRobustCommand{\Ainline}[1]{\begingroup
  \setlength{\fboxsep}{1pt}
  \colorbox{best}{#1}%
\endgroup}

\DeclareRobustCommand{\Binline}[1]{\begingroup
  \setlength{\fboxsep}{1pt}%
  \colorbox{second}{#1}%
\endgroup}

\DeclareRobustCommand{\Cinline}[1]{\begingroup
  \setlength{\fboxsep}{1pt}%
  \colorbox{third}{#1}%
\endgroup}

\centering
\small
\begin{tabular}{l@{\hskip 8pt}c@{\hskip 8pt}|c@{\hskip 12pt}c@{\hskip 8pt}c@{\hskip 8pt}c}
\hline
\textbf{Method} & \textbf{LOD} & \textbf{L1} $\downarrow$ & \textbf{PSNR} $\uparrow$ & \textbf{SSIM} $\uparrow$ & \textbf{LPIPS} $\downarrow$ \\
\hline

                & 0.0  & \A\textbf{0.038} & \A\textbf{28.621} & \A\textbf{0.928} & \A\textbf{0.058} \\
    Ours (full) & 0.5  & \A0.039 & \A28.514 & \A0.927 & \A0.061 \\
                & 1.0  & \A0.041 & \A28.342 & \B0.922 & \A0.072 \\

    \hline
    
             & 0.0 & 0.044 & 27.815   & 0.913   & 0.074 \\
    w/o fmap & 0.5 & 0.044 & 27.864   & 0.914   & 0.074 \\
             & 1.0 & 0.048 & 27.436   & 0.907   & 0.087 \\

    \hline

              & 0.0 & \C0.041    & \C28.295   & 0.913    & 0.063 \\
    fmap (64) & 0.5 & \B0.040    & 28.391   & 0.915    & \C0.065 \\
              & 1.0 & \A0.041    & \C28.321   & \C0.921    & \C0.075 \\

    \hline

               & 0.0 & \C0.041 & 28.289 & \C0.916 & \C0.062 \\
    fmap (128) & 0.5 & \C0.041 & \B28.423 & \C0.917 & \C0.065 \\
               & 1.0 & \B0.042 & 28.191 & 0.912 & \B0.074 \\

    \hline

               & 0.0 & \B0.039 & \B28.614 & \B0.926 & \B0.059 \\
    fmap (256) & 0.5 & \B0.039 & \C28.411 & \B0.924 & \B0.064 \\
               & 1.0 & \A0.041 & \B28.323 & \A0.923 & \B0.074 \\

\hline
\end{tabular}
\caption{\textbf{Quantitative comparison} across different ablation settings and LODs. 
The \textbf{bold} results indicate the globally best score across all methods and LODs. 
Within each LOD setting (0.0, 0.5, 1.0): we color the \Ainline{best} and the \Binline{second best} results.
\textit{w/o fmap} denotes our method without the learnable UV feature map. \textit{fmap (.)} denotes our method where the multi-level UV feature field is replaced by a fixed-resolution learnable feature map, with the value in parentheses indicating the resolution of the feature map.}
\label{tab:ablation}
\end{table}

\begin{figure}[ht]
    \centering
    \includegraphics[width=1.0\linewidth]{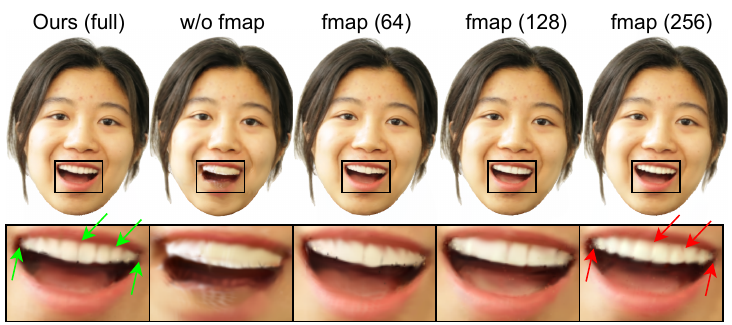}
    \caption{\textbf{Qualitative results of ablation study.} \textit{w/o fmap} denotes our method without the learnable UV feature map. \textit{fmap (.)} denotes our method where the multi-level UV feature field is replaced by a fixed-resolution learnable feature map, with the value in parentheses indicating the resolution of the feature map. \textcolor{red}{Red} arrows indicate artifacts, compared to the corresponding locations marked by \textcolor{green}{green} arrows in our best (full method) result. We recommend zooming in to better inspect the details.}
    \label{fig:ablation_fmap}
    \vspace{-0.5cm}
\end{figure}

We conduct ablation studies on the UV feature field. In the first ablation, we completely remove the UV feature field, meaning no learnable per-Gaussian features (\textit{w/o fmap}). In the second ablation, we replace the multi-level feature field with a single learnable feature map, and test three resolutions: $64 \times 64$, $128 \times 128$, and $256 \times 256$. Self-reenactment experiments are performed on the PointAvatar dataset, with results summarized in Table~\ref{tab:ablation}. We observe that without learnable features, the performance is the worst. When using a single-resolution feature map, higher resolution produces better quality. However, the proposed multi-level feature field (\textit{full}) achieves the best results in most cases. 

In Figure~\ref{fig:ablation_fmap}, we present the qualitative ablation results. Notably, when comparing \textit{fmap (256)} with our best result, we observe more visual artifacts and blurrier sharp edges in \textit{fmap (256)}. We attribute this to the fact that a single feature map must represent all LOD levels, causing the lower LOD representations to negatively affect the higher LOD results.

%% file: sec/5_conclusion.tex
\section{Limitations}
\label{sec:limitations}

Similar to most existing works, our method relies on accurate FLAME tracking to provide reliable 3D-2D alignment, which is essential for maintaining 3D consistency. In addition, we observe that some expression modes appear only under large head poses in the video (e.g., side views). When this happens, the network tends to overfit these rare cases, leading to artifacts when such expression modes occur.

\section{Conclusion}
\label{sec:conclusion}

In this work, we propose ArchitectHead, a framework for creating 3D Gaussian head avatars that supports real-time and continuous adjustment of the level of detail (LOD). To our knowledge, ArchitectHead is the first 3D Gaussian head method to realize continuous LOD control. We parameterize Gaussians in UV feature space, which allows us to control their number by simply adjusting the UV map resolution. A neural decoder then generates the Gaussian attributes, using the LOD as an additional condition. To capture rich local information and balance different LODs while ensuring smooth transitions between them, we introduce a learnable UV latent feature map alongside the UV position map to provide more representative information. We then extend this design to a multi-level latent feature field, which enables weighted resampling across resolutions and improves the balance among varying LODs. Experiments on monocular video datasets show that ArchitectHead achieves state-of-the-art quality. For future work, we plan to extend the framework to multi-view video datasets and improve training efficiency.